# Role of Interestingness Measures in CAR Rule Ordering for Associative Classifier: An Empirical Approach

S.Kannan and R.Bhaskaran

**Abstract** – Associative Classifier is a novel technique which is the integration of Association Rule Mining and Classification. The difficult task in building Associative Classifier model is the selection of relevant rules from a large number of class association rules (CARs). A very popular method of ordering rules for selection is based on confidence, support and antecedent size (CSA). Other methods are based on hybrid orderings in which CSA method is combined with other measures. In the present work, we study the effect of using different interestingness measures of Association rules in CAR rule ordering and selection for associative classifier.

**Key Terms** – associative classifier, class association rule, interestingness measures, rule ordering

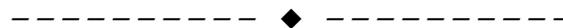

## 1 Introduction

IN the present world of fast communication, massive information is collected from various sources. Storing, processing and maintaining this massive information is certainly a problem at the moment. Data mining is a solution for handling massive heterogeneous information. Association rule mining (ARM), Classification and Clustering are the three major areas of research in data mining. Several well known techniques had been developed in the above areas. Associative Classifier is an effective classifier which is the integration of ARM and Classification. To build an efficient Associative classifier, ranking and selecting more relevant and useful class association rules is the most difficult task. In this work, we study the effect of using various interestingness measures of association rules in ranking and selecting efficient association rules for classification.

This paper is organized as follows. In section 2, we discuss the concept of association rules and class association rules. Section 3 discusses the various steps used in associative classifications such as CAR rules generation, pruning rules, ordering rules, rules selection for classification and prediction of class for new sample. Related works are described in section 4. The dataset used for this work is explained in section 5. Section 6 gives the experimental details, discussion about the results and conclusion.

## 2 Association Rules and Class Association Rules:

Let **D** be a dataset with |**D**| instances or tuples. Let **A**= $\{A_1, A_2, A_3, \ldots, A_m, A_{m+1}\}$ be a set of m+1 distinct attributes and each tuple in **D** is set of attribute values for an attribute set T $\subseteq$ A. For any two disjoint frequent attribute-value subsets $X$ and $Y$ of A, the patterns of the form $X \rightarrow Y$ are called association rules, where $X$ and $Y$ are disjoint sets (ie., $X \cap Y = \emptyset$). For example let Z={ $(A_{i1}, a_{i1})$, $(A_{i2}, a_{i2}), \ldots, (A_{in}, a_{in})$} be the attribute-value set of cardinality n, where $a_{i1}$ is one of the possible value for attribute $A_{i1}$, $a_{i2}$ is one of the possible value for attribute $A_{i2}$, and so on. Frequent attribute-value sets and then association rules can be generated using the popular methods Apriori[2][4], FP-Growth[3][4] or any other well known techniques. The attribute-value sets $X$ and $Y$ are called antecedent and consequent of the association rule respectively.

Class Association Rules (CARs) are the association rules with class label attribute as the only consequent. Let **A**=$\{A_1, A_2, A_3, \ldots, A_m, C\}$ be the m+1 distinct attributes and **C**=$\{c_1, c_2, \ldots, c_t\}$ be the class label attribute with t number of classes. CAR rule is of the form **L** $\rightarrow$ (C,$c_i$), where the pattern **L** is the attribute-value pair from the attribute set {**A** \ **C**} and $c_i$ is the class label value for **C**. Generation of association rules (AR) or class association rules (CARs) is generally controlled by the two measures called support and confidence, which are given below.

*Support* = P(X $\cup$ Y) = P(XY) = (Number of tuples that contains both $X$ and $Y$) / (Total number of tuples in **D**)
*Confidence* = P(Y | X) = P(X $\cup$ Y) / P(X)
     = P(XY) / P(X)

For example consider the weather dataset given in Table-1. Following are the sample class association rules generated for minimum support 10% and minimum confidence 90%. The number of records covered by antecedent and the rule are given in the left side and right side of the implication mark respectively.

Table-1 Weather Dataset

---

- *S.Kannan is with Department of Computer Science, D.D.E., Madurai Kamaraj University, Madurai-625021, Tamil Nadu, India.*
- *R.Bhaskaran is with School of Mathematics, Maduai Kamaraj University, Madurai-625021, Tamil Nadu, India.*

|    | Outlook  | Tempe-rature | Humi-dity | Windy | Play |
|----|----------|--------------|-----------|-------|------|
| 1  | Sunny    | hot          | high      | False | no   |
| 2  | Sunny    | hot          | high      | True  | no   |
| 3  | Overcast | hot          | high      | False | yes  |
| 4  | Rainy    | mild         | high      | False | yes  |
| 5  | Rainy    | cool         | normal    | False | yes  |
| 6  | Rainy    | cool         | normal    | True  | no   |
| 7  | Overcast | cool         | normal    | True  | yes  |
| 8  | Sunny    | mild         | high      | False | no   |
| 9  | Sunny    | cool         | normal    | False | yes  |
| 10 | Rainy    | mild         | normal    | False | Yes  |
| 11 | Sunny    | mild         | normal    | True  | Yes  |
| 12 | Overcast | mild         | high      | True  | Yes  |
| 13 | Overcast | hot          | normal    | False | Yes  |
| 14 | Rainy    | mild         | high      | True  | No   |

**Sample CAR rules:**
1. Outlook=overcast 4 ==> Play=yes 4    conf:(1)
2. Humidity=normal Windy=FALSE 4
            ==> Play=yes 4           conf:(1)
3. Outlook=sunny Humidity=high 3
            ==> Play=no 3            conf:(1)
4. Outlook=rainy Windy=FALSE 3
            ==> Play=yes 3           conf:(1)
5. Outlook=sunny Temperature=hot 2
            ==> Play=no 2            conf:(1)
6. Outlook=sunny Humidity=normal 2
            ==> Play=yes 2           conf:(1)
7. Outlook=overcast Temperature=hot 2
            ==> Play=yes 2           conf:(1)
8. Outlook=overcast Humidity=high 2
            ==> Play=yes 2           conf:(1)
9. Outlook=overcast Humidity=normal 2
            ==> Play=yes 2           conf:(1)
10. Outlook=overcast Windy=FALSE 2
            ==> Play=yes 2           conf:(1)
11. Outlook=overcast Windy=TRUE 2
            ==> Play=yes 2           conf:(1)
12. Outlook=rainy Windy=TRUE 2 ==> Play=no 2
                                     conf:(1)
13. Temperature=mild Humidity=normal 2
            ==> Play=yes 2           conf:(1)
14. Temperature=cool Windy=FALSE 2
            ==> Play=yes 2           conf:(1)
15. Outlook=sunny Temperature=hot Humidity=high 2
            ==> Play=no 2            conf:(1)
16. Outlook=sunny Humidity=high Windy=FALSE 2
            ==> Play=no 2            conf:(1)
17. Outlook=overcast Temperature=hot Windy=FALSE 2
            ==> Play=yes 2           conf:(1)
18. Outlook=rainy Temperature=mild Windy=FALSE 2
            ==> Play=yes 2           conf:(1)
19. Outlook=rainy Humidity=normal Windy=FALSE 2
            ==> Play=yes 2           conf:(1)
20. Temperature=cool Humidity=normal Windy=FALSE
     2 ==> Play=yes 2                conf:(1)

Number of rules grows to several thousands if the support and confidence thresholds are reduced to low. To select interesting rules, different interestingness measures [9] are used to rank the generated rules. Each measure has its own selection characteristics and its own positives and negatives. For detailed discussion, refer [9], [10], [11] and [12]. The Table-2 lists most generally used interestingness measures with their formula for computation.

## 3 Associative Classifications

Associative Classification or Classification Based on Association Rules is a four step process. First Class Association Rules (CARs) for each class are generated using training dataset for the given support and confidence thresholds. Second step is pruning redundant and inefficient CAR rules, using which, will reduce the accuracy of classifier. Third step is the selection of relevant and useful CAR rules for classification. Final step is predicting class label, based on the collection of CAR rules for a new instance which has unknown class label.

CAR rules can be generated using the following well known methods: Apriori based algorithm (as used in CBA[6]), FP-Growth based algorithm (as used in CMAR[7]), FOIL based algorithm (as used in CPAR[8]), Apriori-TFP based algorithm (as used in TFPC (Total From Partial Classification)) and so on.

### 3.1 Ranking and Pruning CAR rules
### 3.1.1 Ranking Rules:

For pruning and ordering rules, the following rule ranking method is generally used in most of associative classifiers. Given two rules R1 and R2 with confidence of the rules as conf(R1) and conf(R2) respectively, support of the rules as sup(R1) and sup(R2) respectively and the number of attribute-value pairs in the left hand side for the two rules as anti-size(R1) and anti-size(R2), R1 is said to be higher ranked rule than R2 if and only if
(1) conf(R1) > conf(R2); or
(2) conf(R1)=conf(R2) but sup(R1) > sup(R2); or
(3) conf(R1)=conf(R2), sup(R1)=sup(R2) but anti-size(R1) < anti-size(R2).
Also a rule R1:**L**$\rightarrow$ c is said to be a general rule with respect to a rule R2:**L**′$\rightarrow$c′, if and only if **L** is a subset of **L**′.

### 3.1.2 Pruning Rules:

The redundant or inefficient rules can be pruned in several ways. One method is pruning more specific and lower confidence rules if there is a general and high confidence rule. Given two rules R1 and R2, if R1 is a general rule w.r.t. R2 and R1 has higher rank than R2, then specific and lower confidence rule R2 will be removed.

Table-2: Interestingness Measures used for Association Rules

| Measure | Formula |
|---|---|
| Support | $P(XY)$ |
| Confidence/Precision | $P(Y \mid X)$ |
| Coverage | $P(X)$ |
| Prevalence | $P(Y)$ |
| Recall / Sensitivity | $P(X \mid Y)$ |
| Specificity-1 | $P(\neg Y \mid \neg X)$ |
| Accuracy | $P(XY) + P(\neg X \neg Y)$ |
| Lift/Interest | $P(Y\mid X)/P(Y)$ or $P(XY)/P(X)P(Y)$ |
| Leverage-1 | $P(Y\mid X) - P(X)P(Y)$ |
| Added Value / Change of Support | $P(Y\mid X) - P(Y)$ |
| Relative Risk | $P(Y\mid X)/P(Y\mid \neg X)$ |
| Jaccard | $P(XY)/(P(X) + P(Y) - P(XY))$ |
| Certainty Factor | $(P(Y\mid X) - P(Y))/(1 - P(Y))$ |
| Odds Ratio | $\{P(XY)P(\neg X \neg Y)\}/\{P(X \neg Y)P(\neg XY)\}$ |
| Yule's Q | $\{P(XY)P(\neg X \neg Y) - P(X \neg Y)P(\neg XY)\}/$ $\{P(XY)P(\neg X \neg Y) + P(X \neg Y)P(\neg XY)\}$ |
| Yule's Y | $\{\sqrt{P(XY)P(\neg X \neg Y)} - \sqrt{P(X \neg Y)P(\neg XY)}\}/$ $\{\sqrt{P(XY)P(\neg X \neg Y)} + \sqrt{P(X \neg Y)P(\neg XY)}\}$ |
| Klosgen | $(\sqrt{P(XY)})(P(Y\mid X) - P(Y))$, $(\sqrt{P(XY)})\max(P(Y\mid X) - P(Y), P(X\mid Y) - P(X))$ |
| Conviction | $(P(X)P(\neg Y)) / P(X \neg Y)$ |
| Collective Strength | $\{(P(XY)+P(\neg Y\mid \neg X)) / (P(X)P(Y)+P(\neg X)P(\neg Y))\} *$ $\{(1-P(X)P(Y)-P(\neg X)P(\neg Y)) / (1-P(XY)-P(\neg Y\mid \neg X))\}$ |
| Laplace Correction | $(N(XY)+1) / (N(X)+2)$ |
| Gini Index | $P(X)*\{P(Y\mid X)^2 + P(\neg Y\mid X)^2\} + P(\neg X)*\{P(Y\mid \neg X)^2 + P(\neg Y\mid \neg X)^2\} - P(Y)^2 - P(\neg Y)^2$ |
| ∅−Coefficient (Linear Correlation Coefficient) | $\{P(XY)-P(X)P(Y)\} / \sqrt{\{P(X)P(Y)P(\neg X)P(\neg Y)\}}$ |
| J-Measure | $P(XY) \log( P(Y\mid X) / P(Y) ) + P(X \neg Y) \log( P(\neg Y\mid X) / P(\neg Y) )$ |
| Piatetsky-Shapiro | $P(XY) - P(X)P(Y)$ |
| Cosine | $P(XY) / \sqrt{(P(X)P(Y))}$ |
| Loevinger | $1 - P(X)P(\neg Y) / P(X \neg Y)$ |
| Information Gain | $\log \{P(XY) / ( P(X)P(Y) )\}$ |
| Sebag-Schoenauer | $P(XY) / P(X \neg Y)$ |
| Least Contradiction | $\{P(XY)-P(X \neg Y)\} / P(Y)$ |
| Odd Multiplier | $\{P(XY)P(\neg Y)\} / \{P(Y)P(X \neg Y)\}$ |
| Example and Counterexample Rate | $1 - \{P(X \neg Y) / P(XY)\}$ |
| Zhang | $\{P(XY)-P(X)P(Y)\} / \max(P(XY)P(\neg Y), P(Y)P(X \neg Y))$ |
| Correlation | $\{P(XY)-P(X)P(Y)\} / \{P(X)P(Y)(1-P(X))(1-P(Y))\}$ |
| Leverage-2 | $P(XY) - P(X)P(Y)$ |
| Coherence | $P(XY) / (P(X)+P(Y)-P(XY))$ |
| Specificity-2 | $P(\neg X \mid \neg Y)$ |
| All Confidence | $\min( P(X\mid Y), P(Y\mid X) )$ |
| Max Confidence | $\max( P(X\mid Y), P(Y\mid X) )$ |
| Kulczynski | $(P(X\mid Y)+P(Y\mid X))/2$ |
| Chi-square $\chi^2$ | $\mid D \mid \sum_{A \in \{X, \neg X\}, B \in \{Y, \neg Y\}} [((P(AB)-P(A)P(B))^2)/(P(A)P(B))]$ |

Pruning can also be performed based on some measure such as chi-square $\chi^2$, Accuracy, WRA (weighted relative accuracy) score and so on. To prune, first the rules are sorted based on one measure, and then rules with measure values lower than the given threshold are pruned. In our work, all possible interestingness measures listed in Table-2 are used in this way to prune CAR rules.

### 3.2 Rule Ordering and Selection of rules for Classification:

#### 3.2.1 Rule Ordering Approaches:

CAR rules can be ordered based on one of the following approaches.

**CSA (Confidence-Support-Antecedent size) Approach:**

In this method, first all rules are sorted in descending order based on confidence, then within the rules those that have same confidence are sorted in descending order based on support and then within the rules those that have same confidence and support values are sorted in ascending order based on the antecedent size.

**ACS (Antecedent size-Confidence-Support) Approach:**

In this method, first all rules are sorted in descending order based on antecedent size, then the rules that have same antecedent size are sorted in descending order based on confidence and then the rules that have same antecedent size and confidence values are sorted in descending order based on the support.

**MCSA (Measure-Confidence-Support-Antecedent size) Approach:**

In this method, first all rules are sorted in descending order based on any one relevant measure M, then within the rules that have same measure values are sorted in descending order based on confidence, then within the rules that have same measure value and confidence are sorted in descending order based on support and then within the rules that have same measure value, confidence and support values are sorted in ascending order based on the antecedent size. In our work, this approach is followed with each one interestingness measures listed in Table-2 as the relevant measure M.

**Single Measure (SM) Approach:**

In this method, all the rules are sorted in descending order based on any one relevant measure M. In our work, this method is used for each interestingness measure listed in Table-2 to find the role of each measure.

**Hybrid Approach:**

In this method, first all the rules are sorted in descending order based on any one measure such as chi-square $\chi^2$, WRA score, Laplace accuracy, leverage and so on. Then top k rules are selected from the sorted rules. These rules are then ordered based on any one of the approaches such as CSA, ACS, CSAFR[14][15] or any one. In our work, this approach is followed by using every one interestingness measure listed in Table-2 in the first level ordering and then using CSA.

#### 3.2.2 Selection of Rules for Classification

Rules are ordered based on any one of the above approaches. From this ordered rules, rules can be selected for classification using any one the following three ways.

1. Best k rules for each case can be selected.
2. First k rules for each class can be selected.
3. All the rules can be used.

In our work, the best k rules method is used for selection of relevant rules for classification. To select best k rules, database rule coverage method as followed in CMAR is used. This is a variation of the method followed in CBA in which k is 1. The method is given below.

**Database Rule Coverage:**
Input :  **R** is set of ordered CAR rules **r:L→c**
          **D** is set of instances **d**
Output: **R**$_{AC}$ is set of selected CAR rules for classification
Algorithm:
  **R**$_{AC}$ = ∅
  Set *Rules-covered-threshold* to 3
                                    /* to fix cover threshold
  Set the *Rules-covered-count* for each instance **d** of **D** to 0
  While (**R** is not empty and **D** is not empty) do
        Remove one CAR rule **r** from the top of **R**
                    /* r is higher ranked rule
        Set the *cover* flag of **r** to *FALSE*
        For each instance **d** in **D** do
          If (left hand side pattern **L** of rule **r** is matched
                                    with instance **d**)   then
                Increment *Rules-covered-count* for **r** by 1
                Set the *Cover* flag of **r** to *TRUE*
          Endif
          If (*Rules-covered-count* of **d** >= *Rules-covered-
                                    threshold*)  then
 /* to ensure each instance d is covered by at least 3 rules
                Remove the covered instance **d** from **D**
          Endif
        Endfor
        If (*Cover* flag of **r** is *TRUE*) then
          Insert the CAR rule **r** into **R**$_{AC}$ at the end
        Endif
  Endwhile

The above algorithm collects rules one by one into **R**$_{AC}$ if the rule covers at least one instance from the training dataset **D**. Also it ensures that each instance of **D** is covered by *Rules-covered-threshold* (here it is three) number of rules. In CBA, this threshold is 1, that is, each instance is covered by at least one rule. Since single rule cover could not predict effectively, multiple rule cover is used to improve efficiency of classification.

### 3.3 Classification based on Association rules:

After ordering and selection, the selected k CAR rules **R**$_{AC}$ can be used to predict class label for new instance for which class label in not known. For the given new instance, the matched rules **R**$_m$ from the selected k rules are collected. If all the matched rules **R**$_m$ have same class label in consequent, then that class label will be assigned to the new instance. Otherwise if the different matched rules have different class labels in consequent, then the class label is decided from scores computed for each class label. For each class, the score is computed using the matched rules for that class. The class label, which has highest score, is assigned to the new instance.

Different scoring methods are used such as Laplace-accuracy score, weighted $\chi^2$ score and so on. In our work, the *weighted* $\chi^2$ score is used to select best class label. The *weighted* $\chi^2$ can be computed as follows.

$$\text{weighted } \chi^2 = \sum [(\chi^2 * \chi^2)/\max \chi^2]$$

where

$\max \chi^2 = (\min\{P(X), P(Y)\} - ((P(X)*P(Y))/|\mathbf{D}|))^2 * |\mathbf{D}| * e$

and $e = \{1/[P(X)*P(Y)]\} + \{1/[P(X)*(|\mathbf{D}|-P(Y))]\} +$
$\{1/[(|\mathbf{D}|-P(X))*P(Y)]\} + \{1/[(|\mathbf{D}|-P(X))*(|\mathbf{D}|-P(Y))]\}$.

## 4 Related works

Agrawal et. al. in [1] initiated the work of association rules, a new area of research. Later Agarwal and Srikant in [2] developed fast algorithm for association rule mining called Apriori. In [3], the authors improved ARM using the method FP-Growth in which candidate item sets are not generated as in Apriori.

Ali et. al. initiated the idea of classification using association rules in [5]. Followed this, Liu et. al. in [6] proved that CBA(Classification Based on Association rules) is an effective classifier than the other traditional classifiers. Apriori based CARM (Class Association Rule Mining) is used in CBA. In CMAR (Classification based on Multiple Asssociation Rules), the authors of [7] used multiple rules to classify data instead of using single rule as in CBA. In CMAR, FP-growth based CARM is used. In some works such as CPAR[8] (Classification based on Predictive Association Rules), the CARM process is integrated with prediction process.

In the works [9], [10], [11] and [12], the authors studied the characteristics of different interestingness measures. Interesting rules can be identified using these measures and other redundant and irrelevant rules can be pruned. The authors Li and Hamilton of [13] discussed the ways of identifying redundant rules and different ways of pruning rules.

The authors in both [14] and [15] discussed different methods of ordering such as CSA (Confidence-Support-Antecedent size), ACS (Antecedent size-Confidence-Support) CSAFR (Confidence-Support-Antecedent size-class distribution Frequency-Row ordering) and other hybrid rule ordering with WRA (Weighted Relative Accuracy), LA (Laplace Accuracy) and $\chi^2$. For rule ordering and selection, even though several methods exist, CSA (Confidence-Support-Antecedent size) is proved to be an efficient method of simple ordering.

## 5. Data Source

Student's dataset of the distance learning program (DLP), contain details about personnel, school studies, seminar classes, materials used, syllabus and other feedbacks. We collected data randomly through questionnaire from UG and PG students of different courses from different seminar centers of the DLP program. After preprocessing, 2000 samples with 1000 samples for each of two classes were used for this analysis. This dataset contains 92 nominal attributes. We randomly selected 50% of samples from each class for training and the remaining 50% samples were used for testing the classifier. Thus 1000 samples were used in training dataset and the remaining 1000 instances were used for test dataset. In this present work, we wanted to only analyze the performance of associative classifier using the number of correctly classified instances, we do not mention here the details of different attributes and the rules generated.

## 6 Results, Discussion and Conclusion

### 6.1 Experimental Results

In our experiment, the student dataset with 1000 instances with two class labels is used as training dataset. Using the training dataset, one lakh class association rules are generated with 10% support threshold and 50% confidence threshold. The interestingness measures are computed as given in Table-2 for all the (1 lakh) CAR rules.

### Classification 1

In this method, first all the (one lakh) rules are sorted based on each measure (for example Accuracy) given in the Table-2. The first 30000 rules are selected and other rules are pruned. These 30000 rules are again sorted based on CSA ordering and these ordered rules are used for selecting relevant CAR rules for classification. Selection of relevant rules is done based on the database rule coverage algorithm as explained above. The third column of Table-3 for that measure denotes the number of selected relevant rules.

Table-3 Accuracy and number of rules used in the associative classification types 1,2 & 3

|  | MprunCSA-30000 | | MCSA-1lakh | | Mprun30000 | |
|---|---|---|---|---|---|---|
|  | Accuracy-1 (out of 1000 instances) | No-rules-1 | Accuracy-2 (out of 1000 instances) | No-rules-2 | Accuracy-3 (out of 1000 instances) | No-rules-3 |
| Preprun15027 | (CSA)757 | (CSA)248 | 762 | 156 |  |  |
| Accuracy | 756 | 265 | 763 | 159 | 763 | 159 |
| AllConfidence | 678 | 260 | 702 | 95 | 702 | 95 |
| Certainity | 756 | 265 | 756 | 265 | 756 | 265 |
| chi2 | 756 | 265 | 746 | 200 | 746 | 200 |
| Coherence | 759 | 267 | 704 | 82 | 704 | 82 |
| Collective-Strength | 758 | 269 | 701 | 84 | 701 | 84 |
| Conviction | 756 | 265 | 756 | 265 | 756 | 265 |
| Correlation | 756 | 265 | 746 | 200 | 746 | 200 |
| Cosine | 748 | 273 | 780 | 84 | 780 | 84 |
| Cover | 678 | 260 | 703 | 79 | 703 | 79 |
| Coverage | 578 | 196 | 576 | 75 | 576 | 75 |
| Change-Support | 756 | 265 | 756 | 265 | 756 | 265 |
| Ex-Cex-rate | 756 | 265 | 756 | 265 | 756 | 265 |
| Gini-Index | 685 | 157 | 565 | 80 | 565 | 80 |
| Infor-Gain | 756 | 265 | 756 | 265 | 756 | 265 |
| Jacard | 759 | 267 | 704 | 82 | 704 | 82 |
| J-Measure | 756 | 265 | 750 | 236 | 750 | 236 |
| Klosgen | 756 | 265 | 751 | 239 | 751 | 239 |
| Kulczynski | 756 | 268 | 790 | 83 | 790 | 83 |
| Laplace-Correction | 756 | 265 | 756 | 265 | 756 | 265 |
| Least-Contrad | 756 | 265 | 763 | 159 | 763 | 159 |
| Leverage-1 | 756 | 265 | 763 | 159 | 763 | 159 |
| Leverage-2 | 756 | 265 | 759 | 262 | 759 | 262 |
| Lift/Interest | 756 | 265 | 756 | 265 | 756 | 265 |
| Linear-Correlation | 756 | 265 | 746 | 200 | 746 | 200 |
| Loevinger | 579 | 130 | 590 | 108 | 585 | 117 |
| Max-Confidence | 756 | 265 | 704 | 195 | 704 | 195 |
| Odd-Multiplier | 756 | 265 | 756 | 265 | 756 | 265 |
| Piatetsky-Shapiro | 756 | 265 | 763 | 159 | 763 | 159 |
| Precision/Confidence | 756 | 265 | 756 | 265 | 756 | 265 |
| Prevalence | 756 | 265 | 756 | 265 | 756 | 265 |
| Qyule | 756 | 265 | 756 | 271 | 756 | 271 |
| Recall/Sensitivity | 678 | 260 | 703 | 79 | 703 | 79 |
| Relative-Risk | 756 | 265 | 745 | 221 | 745 | 221 |
| Sebag-Schoenauer | 756 | 265 | 756 | 265 | 756 | 265 |
| Specificity-1 | 756 | 265 | 761 | 262 | 761 | 262 |
| Specificity-2 | 578 | 196 | 576 | 75 | 576 | 75 |
| Wra-score | 756 | 265 | 763 | 159 | 763 | 159 |
| Yyule | 756 | 265 | 756 | 271 | 756 | 271 |
| Zhang | 756 | 265 | 756 | 265 | 756 | 265 |

Using these selected CAR rules, classification is performed as explained in section 3.3 to predict class label for each instances in test dataset. For the 1000 instances (500 instances in each of two classes), the class labels are predicted. The predicted class label is checked with actual class label of that instance present in the test dataset. The number of correct predictions out of 1000 instances is found and the second columns of the Table-3 show the results. For example, for using accuracy to order and select rules, 265 number of relevant CAR rules are selected and using these, 756 test samples are classified correctly out of 1000 instances as given in the second row of Table-3. Like this, using the test dataset, the number of correct predictions for each other measures (such as AllConfidence, Certainity, chi-square, and so on) is calculated and is shown in Table-3 under MprunCSA30000.

**Classification 2**

In this method of classification, all the one lakh rules are used without pruning. But the rules are ordered based on MCSA method as explained above for every measure M shown in Table-2. These rules in this order are used for selecting relevant CAR rules based on the database rule coverage algorithm. These selected CAR rules are used for classification as above for the test dataset. The number of correct predictions out of 1000 instances is calculated for each measure in Table-2 and these are entered in fourth column of Table-3 under MCSA-1lakh. The number of selected relevant rules is entered in the fifth column of Table-3. For example, for the measure Accuracy, 159 relevant CAR rules were selected and using these, 763 instances out of 1000 were correctly classified. These results are entered in the second row (5[th] and 4[th] columns respectively) of the Table-3.

**Classification 3**

Here, all the (one lakh) rules are first sorted in descending order based on every one measure from Table-2. The first 30000 rules are only used for selecting relevant CAR rules and others are pruned. The 30000 rules in the sorted order are used for selecting relevant CAR rules. The number of selected rules is entered in the seventh column of Table-3. For example, 159 rules are selected for using Accuracy measure. The selected relevant CAR rules are used for classification for the test dataset. The number of correct predictions out of 1000 instances is found for each measure in Table-2 and these are entered in sixth columns of Table-3 under Mprun30000.

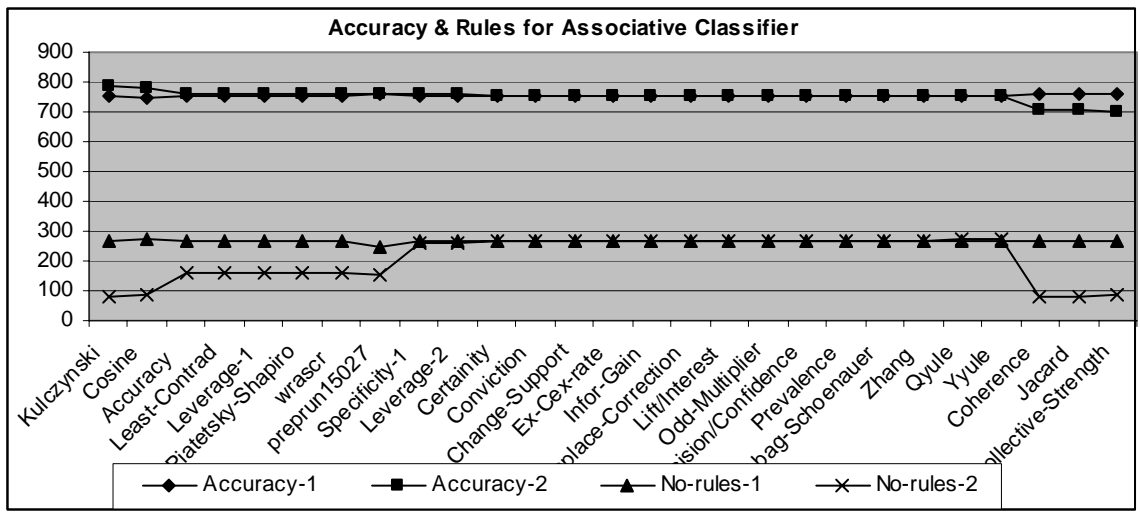

Figure-1 Correct predictions and number of CAR rules for classifications types 1 and 2

## 6.2 Discussion

From the results given in Table-3, it is very clear that the performance of the classification types 2 and 3 are almost same. This means that the classification using ordered rules based on only interestingness measures performs similar to the classification using ordered rules based multi-level ordering (MCSA) using that selected interestingness measure, confidence, support, and antecedent size. In both these classifications, the selected interestingness measure alone plays the role in selecting CAR rules for classification. Consequently for analysis, only the classification results from classification-1 and classification-2 are used. From the classifications 1 and 2, only the better results are plotted in the descending order of accuracy in Figure-1.

From Figure-1, it is observed that the performances of classification-2 using the measures Kulczynski, Cosine, and Accuracy are better than other results. These measures give efficient results with very less number of CAR rules. Also the measures LeastContradiction, Leverage-1, Piatetsky-Shapiro, wrascr, Preprun15027, and Specificity-1 in classification-2 give almost better results with considerably less number of rules in classification-2. The measures Leverage-2, Certainty, Conviction, Change-Support, Ex-Cex-rate, Infor-Gain, Laplace-Correction, Lift/Interest, Odd-Multiplier,

Precision/Confidence, Prevalence, Sebag-Schoenauer, Zhang, Qyule, and Yyule give similar results in classifications 1, 2 and 3. These results are similar to the classification using only CSA rule ordering (given in the 2$^{nd}$ and 3$^{rd}$ columns of first row below the heading). But for the measures Coherence, Jacard, and Collective-Strength, the classification 1 give better results.

## 6.3 Conclusion

For our dataset which is neither dense nor sparse, the measures Kulczynski, Cosine, and Accuracy with classification type 2 yield better results. Also some other measures like LeastContradiction, Leverage-2 play role in classification through rule ordering. From these results, it is concluded that the accuracy of associative classifiers can be improved using appropriate interestingness measures instead of support-confidence framework. Also the measure used for the first level ordering plays an important role not only in selecting relevant CAR rules but also in the accuracy of the associative classifier.

As future work, we intend to study the role of interestingness measures with several standard benchmark datasets. Also we intend to study the role of interestingness measures in predicting class label from the matched rules for a new instance.

**S.Kannan** He completed his M.Sc.(Physics) in 1985 and M.Sc.(Computer science) in 1996 through Madurai Kamaraj University and M.Phil.(Computer Science) through M.S.Univerisity, Thirunelveli. He joined the Madurai Kamaraj University in 1989. At present he is working as Associate Professor. He has presented 5 papers in national and international conferences.

**R.Bhaskaran** He did his M.Sc. from IIT, Chennai in 1973 and obtained his Ph.D. from University of Madras in 1980. He joined the School of Mathematics, Madurai Kamaraj University in 1980. Now he is working as Senior Professor in the school of Mathematics. At present 10 students are working in Data Mining, Image Processing, Software Development, and Character Recognition.